\documentclass[11pt]{article}

\usepackage[preprint]{acl}

\usepackage{times}
\usepackage{latexsym}

\usepackage[T1]{fontenc}

\usepackage[utf8]{inputenc}

\usepackage{microtype}

\usepackage{inconsolata}

\usepackage{graphicx}
\usepackage{booktabs}
\usepackage{float}

\usepackage[most]{tcolorbox}

\newtcolorbox{review}{
  colback=yellow!25,   %
  colframe=yellow!25,  %
  boxrule=0pt,
  breakable,
  enhanced,
  left=2pt, right=2pt, top=2pt, bottom=2pt
}

\title{Linguistic Productivity in Large Language Models: \\Models Coerce, but do not Preempt}

\author{
  \textbf{Claire Bonial\textsuperscript{1}},
  \textbf{Claire Benet Post\textsuperscript{2}},\\
  \textbf{Laura Michaelis\textsuperscript{2}},
  \textbf{Harish Tayyar Madabushi\textsuperscript{3}}
\\
  \textsuperscript{1}DEVCOM Army Research Laboratory, \\
  \textsuperscript{2}University of Colorado Boulder,
  \textsuperscript{3}University of Bath
\\
  \small{
    \textbf{Correspondence:} \href{mailto:claire.n.bonial.civ@army.mil}{claire.n.bonial.civ@army.mil},
    \href{mailto:htm43@bath.ac.uk}{htm43@bath.ac.uk}
  }
}

\begin{document}
\maketitle
\begin{abstract}
Usage-based theories of grammar explain language users' ability to interpret and produce novel combinations of words through two complementary inferential strategies. In cases of \textbf{coercion}, a language user, when encountering a word that is mismatched to its grammatical context, reinterprets that word in a way that resolves the conflict. In cases of \textbf{preemption}, the user infers that an otherwise plausible structure is unavailable or dispreferred because it fails to occur in contexts where it would be expected. Large Language Models are also usage-based, in the sense that the structures of language are learned through exposure to vast amounts of text. Here, we test whether or not the opposing %
forces of coercion and preemption also encourage and constrain linguistic productivity in LLMs. 
We demonstrate across models of various sizes and architectures that larger models recognize  constructional productivity in cases of coercion, even when these involve novel words. 
However, we also show that even the largest models do not replicate the conservative language strategy of preemption for novel words: in particular, they do not appear to draw on negative evidence when making choices among competing grammatical options.  We conclude that coercion is detectable because it is both surprising and semantically motivated: when a word is mismatched to its syntactic context, the word's meaning can be adjusted to fit the context. Preemption has no such signature. There is no record of the lexeme-construction pairings that were avoided---only of those that were produced. This asymmetry helps to explain why models coerce (create mismatched sign combinations) but do not preempt (avoid unconventional yet semantically defensible ones).

\end{abstract}

\section{Introduction}

There is an increasing body of research demonstrating the parallels between language-model construction, based on processing vast amounts of text, and the generalizations that language users deploy in making inferences through 
interaction %
(e.g., \citet{tomasello2006construction,tomasello2009usage}). The success of Large Language Models (LLMs) in a variety of tasks suggests that exposure to the patterns of language in the form of text alone is sufficient to build a functional model of language in which one need not posit grammatical rules a priori.  Usage-based grammatical theories, including Construction Grammar (CxG; e.g., \citet{goldberg1995constructions,croft2001radical,bybee2010language,michaelis2013sign}), likewise hold that grammar is learned from use: repeated exposure routinizes recurrent patterns and supports inference and generalization about the structures of a language \cite{tomasello2003constructing}.  Many such theories treat grammar as a set of conventional solutions that speakers collaboratively develop in order to meet communicative needs they repeatedly face. This approach contrasts with that of Generative Grammar, in which conceptual content comes only from the lexicon, and syntactic rules do no more than determine what symbol sequences function as units for syntactic purposes \cite{kay-michaelis-2012}. In the usage-based approach, grammatical patterns, called holophrases \cite{tomasello2006acquiring}, are seen as symbolic units similar to words---stored and retrieved from memory as individual, whole facts about language.  Grammar thus emerges as conventions do in any recurring social coordination problem: through entrenched improvisations, not as an abstract  rule system \cite{christiansen2022language}.  

Frequency of exposure to a given construction is therefore seen as influencing the course of grammatical development \cite{hoffmann_2022}. Frequency entrenches the selectional properties of grammatical patterns; for example, over time language users come to associate the indefinite article \textit{a} with nouns denoting individuated entities, as in, e.g., \textit{a coin}.  As a result, when such patterns are instantiated by previously unattested combinations of signs, e.g., \textit{a water}, interpreters can resolve the observed semantic conflict by inferring an interpretation of the lexical 'filler' that satisfies the construction's selectional constraints. Thus, \textit{coercion} enables creative extensions of known constructions \cite{michaelis2004type}.  A countervailing force,  \textit{statistical preemption}, or, equivalently, linguistic conservatism, has a constraining effect on productivity in grammar \cite{goldberg2019explain}. For example, statistical preemption explains why speakers do not persist in extending the -\textit{ed} past tense suffix to irregular verbs: in contexts calling for the past tense of \textit{go}, what they reliably hear is the suppletive form \textit{went}, never *\textit{goed}. Thus, speakers attend to negative evidence (what they do not hear in a particular linguistic context) in inferring a grammar.  

Although frequency effects are well established in the psycholinguistic literature, CxG offers few mechanisms to predict which novel formations speakers will accept and which they will reject.\footnote{Licensing-based theories like Sign-Based Construction Grammar (SBCG)  \cite{michaelis2013sign} are the exception; SBCG aspires to complete coverage of the inventory of licit patterns of sign combination.} LLMs provide an unprecedented opportunity to study these forces in vast usage-based models of language.  Here we explore the parallels between human and LLM processing by focusing on two combinatorial phenomena: coercion, which exploits generalizations about which combinations of signs are licit, and statistical preemption, which blocks overgeneralization — the use of sign combinations beyond those one has encountered. \textbf{Our experiments show that models encode constructional categories, including their lexical-class sensitivities, and draw on both to interpret coercive combinations, but do not use the same information to block unattested ones}.

After introducing the theoretical framework (\S\ref{sec:framework}), we review linguistic and psycholinguistic work on productivity, as well as coercion, conceived as a mechanism through which speakers produce novel combinations (\S\ref{sec:coercion}); on preemption, which constrains that productivity (\S\ref{sec:preemption}); and on the interaction between them (\S\ref{ssec:interaction}).  We then present three experiments. The first \textbf{Coercion} experiment tests the models' abilities to interpret coercive combinations of signs (e.g., \textit{I drank the bottle}), wherein the construal of an argument expression is altered to meet the requirements of the linguistic context in which it occurs (e.g., container construed as contents) (\S\ref{sec:method-coercion}).  
Use of coercion is one way in which speakers productively extend the affordances of words. 
The second \textbf{Statistical Preemption} experiment tests the models' ability to constrain productivity by attending to negative evidence: there is a linguistic context in which a structure is semantically viable, and yet never occurs (e.g., \textit{?The asleep cat purred peacefully})\footnote{We precede utterances that may be questionably acceptable to most English speakers with `?'.} (\S\ref{sec:method-preemption}). Thus, statistical preemption curbs the combinatory potential of constructions and thus overgeneralization. The third \textbf{Partial Blocking} experiment explores how models balance these opposing creative and conservative forces in the case of the partial productivity and partial blocking of verbs in Ditransitive and Dative constructions (\S\ref{sec:partial-blocking}).  Our methodology consistently leverages both familiar (English) language and the use of nonce words in order to test for true generalization abilities of models, separate from memorized features of language.  We summarize our results (\S\ref{sec:resuls-coercion}, \S\ref{sec:resuls-preemption}, \S\ref{sec:ditransitive-results}) showing that while models learn constructional categories and lexical sensitivities as to what fits into those categories (positive evidence from priors) to interpret coercive constructions, they do not deploy that information to avoid lexeme-construction combinations unwarranted by the data (negative evidence). Related work is summarized in \S\ref{sec:rel-work} and conclusions offered in \S\ref{sec:conclusions}.

\section{The Usage-Based Theoretical Framework}
\label{sec:framework}

While theories in the tradition of Generative Grammar assume that speakers memorize only a minimal inventory of \textit{signs}, together with a small set of syntactic rules governing the combination of said signs, usage-based theories hold that speakers retain, classify, and organize a rich body of information associated with each token of linguistic experience \cite{bybee2010language}.  Under a CxG approach, these tokens of linguistic experience are represented as signs: pairings of form and meaning that include words, lexemes and phrases \cite{sag2012sign}. These signs are combined by templates, called \textit{constructions}, that may alter the meanings and combinatory potentials of the signs so combined \cite{michaelis2013sign}. 
The form pole of a construction specifies aspects of the pronunciation of the signs combined by the construction \cite{labov1986social,hoffmann_2022}. 
For example, the Comparative-correlative construction licenses a type of bi-clausal sentence in which each clause is introduced by the form \textit{the}, immediately followed by a comparative expression: %
\textit{The higher you fly, the harder you fall.} The meaning pole is the conceptual model associated with the linguistic unit that the construction licenses, which will vary depending upon the speaker.  The meaning pole in CxG is enriched by the use conditions that speakers associate with the sign \cite{hoffmann_2022}.  

As a result of storing various kinds of contextual information with each linguistic exposure, speakers build a grammar beginning with frequently heard linguistic forms associated with frequent experiences \cite{tomasello2006acquiring}. For example, a child learns to associate saying \textit{`Mommy'} with the appearance of this parent.  With further experience, speakers recognize the commonalities between particular linguistic contexts and types of words filling particular slots, such that generalization and information akin to part-of-speech arises \cite{lieven1997lexically}. For example, children may first learn to extend the pattern \textit{Want X} first to more of a certain food or drink, then to more of a desirable item such as a toy \cite{alma991006902879708966}.  The level of abstraction over the learned constructions increases, as speakers first recognize and abstract over commonalities in the slot after \textit{``want,''} then note parallels in the slot before it, then note parallels in broader phrases such that the notion of a verb, verb phrase, and clause arise (speakers deploy this information; but speakers need not be aware of the linguistic labels or meta-linguistic generalization itself) \cite{abbot2006exemplar}.  

In this way, frequency supports the entrenchment of certain structures and patterns of language \cite{bybee2006usage}.  However, it is clear that constructions constrain what their component signs can be, %
and in fact there are some patterns that are semi-productive, wherein the productivity is seemingly idiosyncratic.  For example, the construction that combines a noun with a preceding modifier which we will call the \textit{Attributive Modifier-Head construction}, %
is highly productive: \textit{the red ball; that big building; a fun party.}  A speaker's familiarity with this construction enables interpretation of atypical lexical items within the construction: \textit{a Bob Ross calm}. In other words, the Attributive construction coerces the meaning of the proper noun such that speakers interpret the noun as a modifier.  However, one class of English adjectives resists productive extension to this pattern: \textit{?the asleep cat; ?that aloft balloon; ?an ashamed look}. Notably, there is no clear semantic reason for this incompatibility with the Attributive Modifier-Head construction: a-adjectives do not form a coherent semantic class that would explain their incompatibility with the construction on conceptual grounds. The clash between a-adjectives and the Attributive Modifier-Head construction is rather an artifact of the historical origins of a-adjectives as preposition phrases---an origin story that we cannot assume contemporary English users to know.\footnote{Note that we do discuss a stage-level semantic distinction some have made in \S\ref{sec:preemption}.}

There is, however, a competing viable alternative for expressing that a particular thing has a particular characteristic in English: \textit{the ball is red; the cat is asleep}.  These a-adjectives are attested with some frequency within this Subject-Predicate construction.\footnote{A-adjectives in the Subject-Predicate construction have different information structure: the Attributive Modifier-Head construction presupposes the property ascription (using it to distinguish the head noun's referent from others), whereas Subject-Predicate asserts that the subject NP's referent has the property expressed by the adjective.}  Thus, there is negative evidence blocking the usage of a-adjectives in the Attributive Modifier-Head construction, and positive evidence supporting the usage of a-adjectives in the Subject-Predicate construction.  The fact that speakers do not overgeneralize the Attributive Modifier-Head construction demonstrates that speakers not only attend to the relative frequencies of classes of constructs within a particular class of constructions, but also to negative evidence, or what is never said.

\section{Background}
In the sections to follow, we describe coercion, preemption, and the interaction of these forces.  

\subsection{Coercion}
\label{sec:coercion}
The mechanism of \textit{coercion}, or semantic conflict resolution, was first developed by \citet{pustejovsky1991generative}. Pustejovsky characterizes type coercion as an interpretive operation triggered when a selecting expression is combined with an argument that does not literally instantiate the operator's required semantic type. Rather than rejecting the expression, the interpreter consults the argument's qualia structure to construct a contextually appropriate interpretation that satisfies the selector's type requirements.
For example, the verb \textit{begin} calls for a verbal complement expressing an event that unfolds over time. Thus, \textit{begin the book} is type coercion of the concrete noun \textit{book}, which is thus interpreted in this linguistic context as \textit{begin \textbf{reading} the book.}

In this research, we adopt the definition of coercion given in \citet{michaelis2004type}, who brings a CxG lens to coercion, positing the Override Principle: \\

\begin{quote}
 If a lexical item is semantically incompatible with its morphosyntactic context, the meaning of the lexical item conforms to the meaning of the structure in which it is embedded. \citep[p. 25]{michaelis2004type}. \\
\end{quote}
\noindent Thus, constructional semantics will win out when there is a clash between the lexical class evoked by the construction and the lexical daughter sign actually embedded in the construction.  

Whereas \citet{michaelis2004type} focuses primarily on aspectual coercion, our experiments focus on the coercive process that enables type shifting from container to contents, described in detail in \citet{radden1999towards} as a common metonymic conceptual relationship.  The authors argue that speakers are more interested in the contents of a container than the container itself.  Thus, we commonly see metonymies which target the contents via the container: \textit{This is an excellent bottle} (referring to the wine within the bottle). 

Notice that interpretation and deployment of coercive constructions requires a fair amount of world knowledge that supports conceptual clusters such as containers and contents, but also possessor for possessed, and part-whole metonymy.  This is consistent with a cognitive view of word meaning in which concepts are not defined according to necessary and sufficient conditions, but instead by patterns of metaphorically based semantic extension, 
where ``each metaphor highlights certain aspects of the concept and implicitly hides others'' \cite[p. 201]{lakoff1980metaphorical}.  Thus, coercion facilitates an atypical, but plausible interpretation of a lexical item in a particular constructional context.  Furthermore, coercion is one process supporting the extension of constructions to novel, previously unseen combinations.

\subsection{Statistical Preemption}
\label{sec:preemption}
While psycholinguistic literature has demonstrated that more frequent forms are more likely to serve as the basis for productive generalization \cite{bybee2006spanish}, what prevents speakers from overgeneralizing? Experimental evidence points to  the effect of statistical preemption: 

\begin{quote}
An inference that speakers make from repeatedly hearing form B, when a semantically and pragmatically appropriate expression A could have been used, that B is the appropriate formulation while A is not \citep{suttle2011partial}.
\end{quote} 

\noindent Statistical preemption has been experimentally demonstrated in studies showing that this is how children eschew overgeneralization of certain morphological constructions. Consider again the example of the English past tense ending \textit{``-ed''} on high-frequency irregular verbs: Having repeatedly heard \textit{``went''} in the contexts where \textit{``go-ed''} might be expected is negative, indirect evidence that \textit{``go-ed''} is not appropriate \citep{aronoff1976word,kiparsky1982lexical}. 

The process of statistical preemption at the phrasal level is more complex,  because it is not clear what construction is semantically and pragmatically similar enough to preempt another phrasal form (see \citet{bermel2012corpus,bermel2012morphosyntactic}). 
Nonetheless, compelling experimental evidence for statistical preemption has come from several psycholinguistic studies, including \citet{boyd2012adult} as well as \citet{theakston2004role}. Theakston concludes that entrenchment operates independently from preemption in a task where both children and adults are asked to provide grammaticality ratings for sentences with argument structure errors, involving both high and low-frequency verbs. She finds that frequency plays a strong role: for both children and adults, lower-frequency verbs were found to be more acceptable than high-frequency verbs in the erroneous argument structures. Theakston notes that the adults would certainly be expected to have preempting structures for both high and low-frequency verbs, yet this did not remove the frequency effect observed in participants’ judgments of grammaticality. Thus, we can see that preemption is not an all-or-nothing process, speakers may remain less confident of the range of constructions in which lower-frequency items are used.

Our research draws most directly from the experimental evidence of \citet{boyd2011learning}, who discuss the phenomenon of blocking a-adjectives (e.g., \textit{awake, aware, ashamed, awash}) in the pre-nominal position of the English Attributive Modifier-Head construction (e.g., \textit{?The awake child}).  In Boyd and Goldberg's research, speakers track the fact that although adjectives can generally sit in either the prenominal position of the Attributive Modifier-Head construction (e.g., \textit{the blue bucket}) or in the post-copula position of the Subject-Predicate construction (e.g., \textit{the bucket is blue}), that the phonologically-defined class of English a-adjectives is likely to be observed only in the Subject-Predicate construction (e.g., \textit{the child is awake}). As a result of this \emph{negative evidence}, speakers avoid producing even novel, nonce a-adjectives in the Attributive Modifier-Head construction. 

We note one could interpret the a-adjective restriction as semantically motivated rather than phonological: a-adjectives denote stage-level (temporary) properties, and stage-level properties are independently dispreferred as restrictive prenominal modifiers, since a temporary property is a poorer identifier than an intrinsic one (e.g., \textit{?The asleep cat} vs. \textit{The calico cat}). On this view, the prenominal gap would follow from the semantics of the class rather than from statistical preemption. The design of our stimuli addresses this directly: each a-adjective is paired with a near-synonymous control (asleep/sleeping, afraid/frightened) matched for meaning, so that whatever stage-level influence applies to an a-adjective applies equally to its control. If temporal semantics drove the observed restriction, the two members of each pair would be influenced similarly. 

\subsection{Interaction of Coercion \& Preemption}
\label{ssec:interaction}
We frame the contrast between coercion and preemption as a contrast in the direction of selection: from lexical item to construction or construction to lexical item. In coercion, the speaker reasons from the construction to the lexeme.  Generally, the construction imposes selectional demands on the items that fill its slots, and those demands are semantically motivated. We can give a coherent account of why the Ditransitive selects Transfer semantics, or why a container-denoting noun can stand in for its contents. When a lexeme violates a construction's demands, the speaker resolves the conflict by reinterpreting the lexeme to satisfy the construction, and this reinterpretation is recoverable from the construction's semantics \cite{michaelis2004type}. For example, because the indefinite article is conventionally restricted to count nouns, its use in \textit{a water} prompts the interpreter to infer an additional meaning (e.g., ``an individuated portion of water'') that resolves the mismatch.  

In preemption, the speaker reasons from the lexeme to the construction. The speaker tracks which constructions a word occurs in and which it avoids, and these combinatory facts are idiosyncratic. What a word combines with is not reliably predictable from its meaning. For example, the block on a-adjectives in the pre-nominal Attributive construction is an artifact of their historical origin as preposition phrases, not a fact about their current meaning. On this view, in resolving a coercive conflict, a speaker reasons top-down (semantically motivated constructional selectivity) to enrich the meaning of the mismatched word up to that of the construction (e.g., interpreting \textit{water} in \textit{a water} as a unit of water).  In applying the inferential strategy of preemption, a speaker reasons bottom-up (idiosyncratic lexical combinatorics) to select the conservative formulation of a message (e.g., eschewing the use of  \textit{explain} in the Ditransitive (\textit{?explain me this})) despite the fact that it is more parsimonious than the Dative (\textit{explain this to me}). Note that this complements the positive-evidence and negative-evidence contrast: the two strategies differ not only in whether they rely on what is attested or what is absent, but in whether the constraint they encode is semantically grounded or idiosyncratic.

These forces can also be understood in terms of the \citet{horn1984towards} notion of the division of pragmatic labor, which readily fits with the \citet{goldberg2019explain} discussion of \textit{partial productivity}.  Horn analyzes \textit{partial blocking}, in which a lexicalized form and a productive alternative divide a semantic space rather than one fully blocking the other.  For example, the lexical causative ``stop'' does not fully block the periphrastic ``cause to stop,'' because the periphrastic survives to convey indirect causation. Two forces produce this division: an \textit{economy} force favors the shorter or more conventional form, and a \textit{precision} force favors the form that best conveys the intended meaning. Neither force wins outright, so the result is partial blocking rather than total blocking. Constructional productivity has the same shape: coercion is the productive, expressive force; %
preemption is the conservative check on it. Speakers reach partial productivity because coercion is checked by preemption. 

In order to explore the interaction of the conservative instinct to avoid certain lexical items within certain constructions (the preemptive inference) and the creative drive to express novel or special semantics by deliberately placing an unconventional lexical item within a certain construction (productions involving coercion), we examine the case of the Dative alternation.  This is the titular problem 
presented in \citet{goldberg2019explain}, which, as the title \textit{Explain me this} suggests, includes an examination of the partial productivity and partial blocking of the Ditransitive construction (sometimes also referred to as the Double-Object (DO) construction) and Dative constructions, also previously explored in \citet{goldberg2011corpus}.  This research, as well as that of dozens of other linguists,  focuses on explaining the fact that some `alternating' verbs can felicitously combine with both constructions---for example, both the motion verb \textit{kick} and the transfer verb \textit{give}: \\

\noindent \textbf{Ditransitive/DO}: \textit{I gave/kicked him the ball.} \\
\textbf{Dative}: \textit{I gave/kicked the ball to him.} \\

\noindent However, there are also sets of verbs that are found almost exclusively in corpora in the Dative, but not the Ditransitive variant (for example, ``push:'' \textit{Push the box to me; ?Push me the box}).  
Explanations for why certain verbs participate in this alternation while others do not range from morphosyntactic and diachronic (verbs with morphosyntactic properties of Latinate verbs are blocked from the Ditransitive) to semantic, with the semantic criteria ranging over the verbs (transfer verbs vs. caused-motion verbs) and the arguments (Recipients/Goals must be animate), to information structure properties (Ditransitive Recipients must be short and pronominal, while the Themes are indefinite, complex noun phrases).  It is not within the purview of this work to determine \textbf{why} this partial blocking occurs; instead, we embrace the view of \citet{gould2018match}, who propose the ``hybrid model'' combining (a) discourse-pragmatic and production factors, such as pronominality and grammatical weight, which affect syntactic distributions and are constrained by semantic properties of arguments and verbs, and (b) constructional meaning, which may enrich or alter the semantic representations of verbs \cite[239]{gould2018match}. \citet{gould2018match} experimentally demonstrate that speakers exhibit sensorimotor simulation distinctions between caused-possession (what we will refer to as \textit{Transfer} verbs here), and caused-motion verbs.  

We are specifically interested in whether or not models demonstrate similar conservative and creative strategies when presented with novel verbs: how likely are models, compared to people (children), to accept a novel verb in a Ditransitive or Dative variant? We leverage the findings of \citet{gropen1989learnability}, who analyze both corpus evidence and spontaneous production of children with respect to the dative alternation.  Their findings support a similar hybrid view of a combination of factors guiding what they term ``criteria-governed productivity'' where those criteria range in the ways described above, from morphosyntactic to semantic.  What we take from this study is the finding that there is a certain subclass of motion verbs, \textit{Manner of Accompanied Motion} (e.g., \textit{push, carry, pull}) that never appear in the  Ditransitive variant.  In contrast, all \textit{Giving/Transfer} verbs (e.g., \textit{give, sell, hand}) readily appear in both formulations and are the first to arise in children's production of the Ditransitive.    

Thus, we focus on model judgments of the naturalness of either Transfer verbs or Manner of Accompanied Motion verbs in the Dative and Ditransitive alternations.  When encountering a novel verb that clearly highlights Manner of Accompanied Motion semantics (established through context sentences where the novel verb replaces real verbs such as ``pull / drag / tow''), the conservative, preemptive force should lead speakers to prefer the novel verb in the alternative, safer formulation of the Dative.  Thus,  \textit{He pulled/bripped a trailer to me} is preferred over \textit{He pulled/bripped me a trailer}.  We test whether models, like children (as demonstrated in \citet{gropen1989learnability}), extend Ditransitive blocking patterns to that novel verb.\footnote{We wrestled with whether or not this blocking should be treated as a second case of statistical preemption.  We ultimately decided that we do \textbf{not} consider this blocking to be equivalent to the kind of statistical preemption described with respect to a-adjectives because this partial blocking may be related to semantic incongruity between Manner of Accompanied Motion verbs and the Ditransitive, as opposed to being an idiosyncratic fact about the language that must be learned through negative evidence.}  Our experimental results demonstrate that models, in contrast with children, carry the creative, productive force but lack the conservative check on novel items, so productivity runs unchecked.

\section{Experimental Methodology}
\label{sec:method}
We aim to test LLM abilities with respect to linguistic generalization and the inferences required of both coercion and preemption: when to accept and make sense of a novel combination on the one hand, and when to eschew the use of a novel combination on the other hand.  We draw upon the research summarized in the previous section to set up experiments in which we compare model performance to human preferences in selecting between 1) type-shifting coercive constructions, 2)  alternative formulations of a-adjectives, and 3) Transfer or Manner of Accompanied Motion verbs in Dative/Ditransitive constructions.  In all three experiments, testing for true generalization requires an evaluation methodology that precludes model memorization of data masquerading as generalization of a pattern. Thus, in all cases, we test novel nonce words. This ensures that performance on the nonce test items cannot be accounted for via memorization.  

\subsection{Models and Measures}
\label{sec:models}
We test across the models listed in Table~\ref{tab:models}, providing a cross-family and cross-paradigm panel of models for robustness of our results. 

\begin{table*}[!h]
\centering
\begin{tabular}{@{}lllll@{}}
\toprule
Model & Approx params & Family & Post-training & Paradigm \\
\midrule
gpt-4o-mini           & sub-frontier          & OpenAI   & instruct      & forced choice$^{\dagger}$ \\
gpt-5.4               & frontier              & OpenAI   & instruct      & forced choice \\
deepseek-v4-flash     & 284B MoE (13B active) & DeepSeek & instruct      & forced choice \\
Llama-3.1-8B base     & 8B                    & Meta     & pretrain only & surprisal \\
Llama-3.1-8B-Instruct & 8B                    & Meta     & instruct      & forced choice \\
Qwen2.5-7B-Instruct   & 7B                    & Alibaba  & instruct      & forced choice \\
Qwen3-32B             & 32B                   & Alibaba  & instruct      & surprisal (8-bit) \\
\bottomrule
\end{tabular}
\caption{Models evaluated. $^{\dagger}$Scored two ways: the discrete A/B pick and the answer-token log probability.}
\label{tab:models}
\end{table*}

We use two scoring paradigms. For the base model (Llama-3.1-8B base), we take autoregressive surprisal on each candidate sentence conditioned on the establishment context and select the lower-surprisal option; there is no A/B swap, since surprisal is deterministic per option. For the instruct models, we present both options as A and B and let the model pick one, counterbalancing AB and BA orderings to absorb slot bias. The two paradigms read different things off the same item: surprisal reads lexical statistics directly, forced-choice reads metalinguistic acceptability. Running both lets us triangulate the construction-level signal, and any divergence between the two is itself a finding rather than noise. We  test both forced-choice and surprisal on gpt-4o-mini, which allows us to check for cross method agreement. Finally, we score Qwen3-32B, a post-trained model, by surprisal: it is hosted locally with full access to token probabilities, and it provides the larger-scale comparison point using surprisal.

We note that surprisal, log probability, and perplexity are closely related transformations of the same underlying quantity. Surprisal is the negative log probability a model assigns to a token; perplexity is the exponential of the mean surprisal. ``Log probability'' and ``perplexity'' just describe how the surprisal was obtained, not a different measure.

In these experiments, we adopt the term \textit{surprisal}. However we obtain this measure via perplexity (surprisal taken from the model run as a language model over the stimulus) for the base decoder (Llama-3.1-8B base) and Qwen3-32B.
We obtain this via log probabilities (surprisal taken from the model's answer-token log probabilities — a soft forced choice) for the GPT-4o-mini.  All other model results are forced choice (the model's discrete A/B pick), not a surprisal measure.

\subsection{Experiment 1: Coercion}
\label{sec:method-coercion}
The coercion that we focus on is coercive type-shifting involving nouns that can denote either a container or its contents. As described previously, linguistic literature posits that a reference to the container can stand in for the contents. Thus, upon hearing \textit{This is an excellent bottle!} a listener may construe from the linguistic context and commonsense information that the speaker is complimenting the contents of the bottle, not the bottle itself. Our experimental setup tests whether or not models are attuned to both the broader linguistic context and commonsense knowledge such that either model surprisal (in base decoder models) or forced-choice reflects the forced-choice acceptability judgments of native English speakers.  

Leveraging linguistic studies as the basis, we synthesize a dataset of 60 paired sentences with 
novel, nonce container or contents nouns. Preceding each target coercive sentence, there are 4 context-setting sentences that highlight either the container usage of the coerced noun or the contents usage. For example, both models and annotators were asked to choose either Option A or B as sounding more natural: 
\begin{enumerate}
\item Option A: Jim rinsed the glempkit under the tap. He placed the glempkit on the stove. He corked the glempkit again. He lifted the glempkit by the rim. \textbf{Then he sipped the glempkit.}
\item Option B: Jim blended the tropmir with the milk. He squirted the tropmir from the bottle. He dissolved the tropmir in the water. He spread the tropmir across the dough. \textbf{Then he chipped the tropmir.} 
\end{enumerate}

\noindent While the Option A context-setting sentences force an interpretation of the nonce word as some type of container, those of Option B force the interpretation of another nonce word as a substance. The final, bold-faced sentence provides the coercive context. For Option A, what is clearly a container is coerced into a substance reading in the context of the verb \textit{``sip.''} For Option B, what is clearly a substance is coerced into a container reading in the context of the verb \textit{``chipped.''} In this case, Option A represents the coercion of the container as contents, and Option B represents the coercion of the contents as container (which option is presented as A or B is randomized in testing).  Human annotation and the linguistic literature point to Option A (Container as Contents) as the more natural choice.  

The familiar set comprises 60 pairs across six verb pairings. Across the set of 60 sentences with nonce nouns, we compare model and annotator choice of which option is more natural. Annotators were two native English speakers with graduate-level education.  For base decoder models where we can measure surprisal (through perplexity), we consider the option with lower surprisal to be the more natural option. 

\subsection{Experiment 2: Statistical Preemption}
\label{sec:method-preemption}

For the a-adjective stimuli, we refine and extend the test items of \citet{boyd2011learning} to collect a set of 40 test items where a-adjectives are paired with near synonyms that are not a-adjectives (e.g., \textit{asleep, sleeping}). Both the a-adjective and the synonym are presented in the Attributive Modifier-Head construction and the Subject-Predicate construction within a relative clause. In the nonce variant of the dataset, the a-adjectives and synonymous counterparts (that do not begin with ``a'') are replaced with novel nonce words carefully selected to preserve the a-adjective property while not invoking real words. For example, consider the following set of nonce test items which replace the familiar words \textit{asleep, sleeping}: \\

\noindent \textbf{A-adjective; attributive}: \textit{The asneld cat purred softly on the cushion}.\footnote{We emphasize that there are other contexts in which the Attributive construction is a very attractive option, e.g., in a context of contrast: \textit{not the black cat, the \textsc{asneld} cat}. In future work, we plan to conduct the same experiments using these kinds of attractive contexts to see if it outweighs the force of the preemption.}\\
\textbf{A-adjective; predicative}: \textit{The cat that was asneld purred softly on the cushion}. \\
\textbf{Synonym; attributive}: \textit{The blelming cat purred softly on the cushion}. \\
\textbf{Synonym; predicative}: \textit{The cat that was blelming purred softly on the cushion}.\\

We rely on previous human productions from \citet{boyd2011learning} to posit that a-adjectives are dispreferred in the pre-nominal Attributive Modifier Head construction, but acceptable in the Subject-Predicate construction, while other synonymous adjectives are acceptable in either construction.\footnote{Because we acknowledge that model forced choice is not directly comparable to human production, we compare the direction and magnitude of the class contrast rather than the rates themselves. Notably, our denominator is also different when comparing results, as human preferences are across 20 items, while our model forced choice is across 40 items.}  Specifically, we parallel Experiment 1 of \citet{boyd2011learning}, which demonstrated that people extend the preemptive pattern of blocking a-adjectives in the Attributive construction to novel a-adjectives, even when those novel a-adjectives had never been witnessed in an exposure phase of the experiment.\footnote{Experiment 2 of \citet{boyd2011learning} introduces the novel a-adjective in the alternative, Subject-Predicate construction, and shows that the blocking is stronger when people are exposed to even one or two instances of the alternative, preempting pattern.  In our experiment, we do not introduce any novel a-adjectives in alternative formulations; thus, we compare model performance to Experiment 1 results.}  We present the models with the familiar and nonce test items, and measure surprisal for base decoder models and collect forced-choice responses from other models. We compare observed production preferences to the forced-choice responses and investigate whether or not surprisal is higher for the test items that would be preempted by people.  

\subsection{Experiment 3: Partial Blocking}
\label{sec:partial-blocking}
To explore the partially productive, partially blocked patterns of Transfer and Manner of Accompanied Motion verbs in the  Ditransitive/Dative alternation, we synthesize a dataset of 40 test items comprised of familiar, English words and a parallel dataset of 40 test items where nonce words stand in for a target verb.  We present 4 context-setting sentences prior to the target test sentence that instantiates the familiar or novel verb in the Ditransitive and the Dative constructions. For example, consider the following set of nonce test items which replace the Transfer verb \textit{``sell''} and the Manner of Accompanied Motion verb  \textit{``transport''} respectively: \\

\begin{small}
\noindent \textbf{Establishment context for novel Transfer verb}: \textit{The baker blesped a fresh loaf in exchange for two pound coins this morning. Every Saturday, Maria the market vendor blesps a small bunch of flowers for three pounds. The butcher blesped the cut of meat, and the regular customer carried it home for supper. The fishmonger blesped a parcel of cod for Mrs Patel, who comes in every Friday.}\\
\textbf{Dative}: \textit{Sarah blesped the document to her sister.}\\
\textbf{Ditransitive}: \textit{Sarah blesped her sister the document.}\footnote{In ongoing work, we have replicated these experiments using stimuli with information structure properties that are more compatible with the Ditransitive (i.e. pronoun as the Recipient, and a complex, indefinite noun phrase as the Theme). In these cases, therefore, the Ditransitive is the more attractive option with respect to information structure properties. Our preliminary findings show that these changes do not alter the results we present here; we will present these findings in detail in future work.}\\
\textbf{Establishment context for novel Manner of Accompanied Motion verb}: \textit{The conveyor belt brulmed the boxes across the warehouse floor.  At dawn, the forklift brulms the pallets through the narrow loading bay. The freight train brulmed the steel girders down the long track and finally reached the depot. Later, a tractor brulmed the logs along the muddy path toward the timber yard.}\\
\textbf{Dative}: \textit{Sarah brulmed the document to her sister}\\
\textbf{Ditransitive}: \textit{Sarah brulmed her sister the document}\\
\end{small}

\noindent For the novel Transfer verb, we do not expect either the Dative or the Ditransitive to be strongly preferred or dispreferred, as Transfer verbs commonly appear in either construction (as shown across a variety of research, including \citet{gropen1989learnability,hovav2008english} and \citet{gould2018match}).  For the novel Manner of Accompanied Motion verb, we expect a preference for the Dative, as these verbs will not have been witnessed in the Ditransitive, but would be commonly observed in the Dative.  

We present each test item and compare forced-choice responses as to whether the Dative or Ditransitive is more natural sounding, and we also compare whether or not higher surprisal aligns with expectations from existing corpus literature \cite{gropen1989learnability}.  Specifically, based on the existing experimental and corpus evidence, if speakers extend blocking patterns to other class members, then we would expect the verbs highlighted by context to be Transfer verbs to be acceptable in the Ditransitive, while verbs highlighted by the context to be Manner of Accompanied Motion verbs to be preferred in the Dative.  This preference reflects conservatism in language: if various motion verbs have been witnessed regularly in the formulation of the Dative, and only rarely or never witnessed in the Ditransitive, then the conservative choice is to prefer the Dative.  

\begin{table*}[!h]
\centering
\begin{tabular}{lllr}
\toprule
Tier & Model & Scoring & \shortstack{Familiar\\Container$\rightarrow$ Contents} \\
\midrule
Closed --- large       & GPT-5.4               & forced choice        & 90\% \\
\addlinespace
Closed --- mid         & GPT-4o-mini           & forced choice        & 99\% \\
Closed --- mid         & GPT-4o-mini           & forced choice (logprob)              & 99\% \\
\addlinespace
Open --- MoE (13B active)       & DeepSeek V4-Flash     & forced choice        & 66\% \\
\addlinespace
Open --- large ($\sim$32B) & Qwen3-32B (instruct)    & per-token surprisal  & 87\% \\
\addlinespace
Open --- small (7--8B) & Qwen2.5-7B-Instruct   & forced choice        & 82\% \\
Open --- small (7--8B) & Llama-3.1-8B base     & per-token surprisal  & 80\% \\
Open --- small (7--8B) & Llama-3.1-8B-Instruct & forced choice        & 74\% \\
\midrule
---                    & Expected Preference      & ---        & 100\% \\
\bottomrule
\end{tabular}
\caption{Rate at which models select the Container-coercing-Contents option for stimuli built from familiar English words, compared against the expected preference for Container-coercing-Contents based on theory. Every model is above chance, so the construction is recovered across vendors, scales, and scoring methods. Closed models are tiered by relative size within their family (5.4 $>$ 4o-mini $>$ V4-Flash); exact parameter counts are undisclosed. Open models are tiered by known parameter count.}
\label{tab:exp1aresults}
\end{table*}

\begin{table*}[!ht]
\centering
\begin{tabular}{lllr}
\toprule
Tier & Model & Scoring & \shortstack{Novel \\ Container$\rightarrow$ Contents} \\
\midrule
Closed --- large       & GPT-5.4               & forced choice       & 75\% \\
\addlinespace
Closed --- mid         & GPT-4o-mini           & forced choice       & 60\% \\
Closed --- mid         & GPT-4o-mini           & forced choice (logprob)             & 59\% \\
\addlinespace
Open --- MoE (13B active)       & DeepSeek V4-Flash     & forced choice       & 55\% \\
\addlinespace
Open --- large ($\sim$32B) & Qwen3-32B  (instruct)    & per-token surprisal & 85\% \\
\addlinespace
Open --- small (7--8B) & Qwen2.5-7B-Instruct   & forced choice       & 49\% \\
Open --- small (7--8B) & Llama-3.1-8B base     & per-token surprisal & 65\% \\
Open --- small (7--8B) & Llama-3.1-8B-Instruct & forced choice       & 53\% \\
\midrule
Human annotation       & ---                   & forced choice       & 94\% \\
\bottomrule
\end{tabular}
\caption{Results summary for stimuli testing the interpretation of coercive constructions; the final column compares the percentage of Container coercing Contents reading to the percentage of human preference for this choice at 94\%.}
\label{tab:exp1results}
\end{table*}

\section {Results}

\subsection{Experiment 1: Coercion}
\label{sec:resuls-coercion}
Our first coercion experiment examines model responses or surprisal across test items that involve familiar, English words for either containers or contents (e.g., \textit{Mary drained the bottle} (Container-coerces-Contents) vs. \textit{Mary broke the wine} (Contents-coerces-Container)).  We compare model preference to the expected preference drawn from the linguistic literature (described in \S\ref{sec:coercion}), which calls for a trend towards Container-coerces-Contents.  We also conduct rounds of human evaluation of these stimuli as we curated them, such that there is no overlap with meaningful idiomatic expressions such as \textit{Mary smashed the wine}, which can denote drinking the wine quickly in some English dialects. 

Results, summarized in Table~\ref{tab:exp1aresults}, demonstrate that all models are above chance in selecting the expected Container-coerces-Contents option.  GPT-4o-mini approaches the expected rate, at 99\%.  We conclude that all models recover the constructional semantics despite the mismatched container noun within the construction that generally combines with a drinkable contents noun.  

Our second coercion experiment compares model responses or surprisal across test items that first provide a context establishing or highlighting a nonce noun's status as either a container or contents. The target sentence then forces a choice between a construction coercing a contents reading of something established as a container, or a container reading of something established as a substance/contents.

The human forced-choice annotation demonstrates that coercion is readily available in the direction of coercing the container to refer to the contents, as expected given relevant linguistic literature.  Specifically, pooling results across two annotators, people prefer this coercion option in 94\% of the test items.\footnote{Annotator 1 vs expected result 59/60=98.3\%; Annotator 2 vs. expected 54/60=90.0\%. Raw inter-annotator agreement is 88.3\%. Cohen's $\kappa$ is uninformative at these base rates (prevalence paradox \cite{feinstein1990high,cicchetti1990high}). Thus, we report PABAK (prevalence adjusted and bias-adjusted $\kappa$) of 0.77 \cite{byrt1993bias}.} Table~\ref{tab:exp1results} summarizes comparative results across models and human preference.

The largest models approach human-like preferences: we measure Qwen3-32B preference based on per-token surprisal and find higher surprisal for the Contents coercing Container option, indicating that the Container coercing Contents option is less surprising and more expected 85\% of the time. 
Notably, we see near identical results for GPT-4o-mini, where we measure via both the discrete pick and answer-token log probabilities. DeepSeek V4-Flash, despite its frontier-scale total parameter count, activates only 13B parameters per token and its performance is similar to the smaller open-weight models, consistent with activated capacity rather than total size driving generalization.  

Importantly, our result shows that even when faced with nonce words, models, like people, are able to understand from prior context the semantics of the target lexical item, and furthermore, interpret the coerced semantics such that models reflect a human-like preference for the Container as Contents coercion in comparison to Contents as Container.  It is worth emphasizing that this requires models to interpret novel lexical items on the fly from context, which in turn requires tapping into commonsense information about what kinds of things have substance-like properties and which have container-like properties. Increased model scale helps models generalize, such that novel instantiations of constructions can be interpreted in a human-like fashion when faced with coercive combinations of certain lexical items in conflicting constructional contexts.  

\subsection{Experiment 2: Preemption}
\label{sec:resuls-preemption}
We first compare LLM results against human production data from \citet{boyd2011learning} of familiar and nonce a-adjectives in either the Attributive Modifier-Head construction or the post-copula Subject-Predicate construction.  

Table~\ref{tab:exp2aresults} summarizes model choice in comparison to human preference (``B\&G humans'' referring to the human subjects of \citet{boyd2011learning}) for familiar, English words. Human preference is strongly for non-a-adjectives in the  Attributive Modifier-Head position: people prefer the a-adjective at a rate of 20\% in comparison to the non-a-adjective in the same position at a rate of 78\%. This provides a difference or gap of 58 percentage points; we use this gap as a measure of the strength of preference for people.

When compared to model results, we see that all of the instruction-tuned models demonstrate the same statistical preemption blocking of the a-adjective and preference for the non-a-adjective in the Attributive construction. 

Only Qwen3-32B and Llama-3.1-8B base are scored by surprisal \textit{argmin} (i.e. pick the lower-surprisal form per pair). The Subject-Predicate form has lower surprisal on nearly every pair (79/80) for both a-adjectives and non-a-adjectives as the relative-clause frame adds left context before the adjective.  As a result, \textit{argmin} floors the Attributive Modifier-Head preference rate to ~0 for a-adjectives and controls alike, collapsing the gap to ~0. This is a measurement floor, not a failed restriction. The whole measurement lives in the magnitude of the surprisal difference (in \textit{nats}, or \textit{bits}); \textit{argmin} throws that away by reducing it to a binary pick. Therefore, an alternative way to score a base decoder is the graded Class × Position interaction, which is clearly positive: a-adjectives carry a larger Attributive penalty than the controls (Qwen3-32B +2.81 nats, Llama-8B +3.07 nats). This confirms that these  models mirror the preemptive behaviors of people.  

Overall, our results demonstrate that for familiar English test items, the blocking via statistical preemption holds for all models: the a-adjectives are strongly dispreferred in the pre-nominal position of the Attributive Modifier-Head construction.  
\begin{table*}[!h]
\centering
\resizebox{\textwidth}{!}{%
\begin{tabular}{lllrrr}
\toprule
Tier & Model & Scoring & \shortstack{Familiar\\\emph{a-}adj\\Attributive} & \shortstack{Familiar\\non-\emph{a-}adj\\Attributive} & Gap \\
\midrule
Closed --- large       & GPT-5.4               & forced choice    & 22.5\% & 95.0\% & $-72.5$ \\
\addlinespace
Closed --- mid         & GPT-4o-mini           & forced choice    & 42.5\% & 97.5\% & $-55.0$ \\
Closed --- mid         & GPT-4o-mini           & forced choice (logprob)        & 42.5\% & 97.5\% & $-55.0$ \\
\addlinespace
Open --- MoE (13B active)       & DeepSeek V4-Flash     & forced choice    & 62.5\% & 97.5\% & $-35.0$ \\
\addlinespace
Open --- large ($\sim$32B) & Qwen3-32B         & surprisal argmin & 0.0\%  & 0.0\%  & \phantom{$-$}$0.0$ \\
\addlinespace
Open --- small (7--8B) & Qwen2.5-7B-Instruct   & forced choice    & 17.5\% & 85.0\% & $-67.5$ \\
Open --- small (7--8B) & Llama-3.1-8B base     & surprisal argmin & 0.0\%  & 5.0\%  & $-5.0$ \\
Open --- small (7--8B) & Llama-3.1-8B-Instruct & forced choice    & 45.0\% & 72.5\% & $-27.5$ \\
\midrule
---                    & B\&G humans           & production       & 20.0\% & 78.0\% & $-58.0$ \\
\bottomrule
\end{tabular}
}
\caption{Attributive construction preference rates for \emph{a-}adjectives vs.\ non-\emph{a-}adjectives, with the gap (\emph{a-}adj $-$ non-\emph{a}). Negative values indicate the expected restriction on \emph{a-}adjectives in Attributive construction.}
\label{tab:exp2aresults}
\end{table*}

Table~\ref{tab:exp2bresults} summarizes model choice in comparison to humans for the nonce a-adjectives and nonce non-a-adjectives in the Attributive Modifier-Head construction. Note that the human dispreference for novel a-adjectives in the Attributive construction is not as strong as it was for familiar a-adjectives.  Nonetheless, the preference is there, which \citet{boyd2011learning} suggest is an indication that preemption is extended to the class of a-adjectives even when subjects have not been exposed to that specific word in an alternative, preempting alternation: ``speakers show a significant dispreference for the attributive use of novel a-adjectives. This behavior is consistent with the existence in the grammar of an abstract a-adjective representation'' \cite[p.71]{boyd2011learning}. Models, on the other hand, do not reflect this extension of the preempting pattern to novel items in the class of a-adjectives: no model produces a similar gap, and in fact the gaps collapse to roughly zero or hold in the wrong direction (i.e. models show no preference for a-adjectives or non-a-adjectives in either position, or prefer the a-adjectives in the attributive).  Thus, our findings demonstrate that while models can generalize preemption patterns for familiar words, models cannot deploy this generalization in the face of novel words.  
\begin{table*}[!h]
\centering
\resizebox{\textwidth}{!}{%
\begin{tabular}{lllrrr}
\toprule
Tier & Model & Scoring & \shortstack{Novel\\\emph{a-}adj\\Attributive} & \shortstack{Novel\\non-\emph{a-}adj\\Attributive} & Gap \\
\midrule
Closed --- large       & GPT-5.4               & forced choice    & 75.0\% & 65.0\% & \phantom{$-$}$+10.0$ \\
\addlinespace
Closed --- mid         & GPT-4o-mini           & forced choice    & 80.0\% & 85.0\% & $-5.0$ \\
Closed --- mid         & GPT-4o-mini           & forced choice (logprob)        & 80.0\% & 85.0\% & $-5.0$ \\
\addlinespace
Open --- MoE (13B active)       & DeepSeek V4-Flash     & forced choice    & 70.0\% & 57.5\% & \phantom{$-$}$+12.5$ \\
\addlinespace
Open --- large ($\sim$32B) & Qwen3-32B         & surprisal argmin & 75.0\% & 30.0\% & \phantom{$-$}$+45.0$ \\
\addlinespace
Open --- small (7--8B) & Qwen2.5-7B-Instruct   & forced choice    & 42.5\% & 50.0\% & $-7.5$ \\
Open --- small (7--8B) & Llama-3.1-8B base     & surprisal argmin & 75.0\% & 20.0\% & \phantom{$-$}$+55.0$ \\
Open --- small (7--8B) & Llama-3.1-8B-Instruct & forced choice    & 35.0\% & 42.5\% & $-7.5$ \\
\midrule
---                    & B\&G humans           & production       & 62.0\% & 88.0\% & $-26.0$ \\
\bottomrule
\end{tabular}
}
\caption{Attributive construction preference rates for nonce \emph{a-}adjectives vs. nonce non-\emph{a-}adjectives, with the gap (\emph{a-}adj $-$ non-\emph{a}). Positive values indicate that the model places \emph{a-}adjectives in Attributive construction \emph{more} often than non-\emph{a}-adjectives---the opposite of the human pattern.}
\label{tab:exp2bresults}
\end{table*}

We note that a potential confound may be that while humans recognize the a-adjective character of the nonce adjectives, and therefore extend preemption patterns to the novel adjectives, models do not recognize this feature. We therefore test whether or not models can accurately identify the novel a-adjectives in comparison to other nonce adjectives.  We find that models are able to identify the distinct adjective classes near 100\% of the time (full results reported in Appendix A).  Every frontier model identifies the class accurately from form. Thus, the category is recoverable from text: the model knows \textit{asneld} patterns with \textit{asleep} and \textit{blelming} does not. 

The novel forced-choice failures are about deployment, not identification: GPT-5.4's collapse (showing a clear opposite preference for a-adjectives in the pre-nominal, Attributive construction) was not class-blindness, as we know that when asked directly, it classifies the same words correctly.  We conclude that the a-adjective category is extractable and identifiable to models, but preemption does not follow: the models do not apply the category restriction productively to novel a-adjectives without being told to do so.

\subsection{Experiment 3: Partial Blocking}
\label{sec:ditransitive-results}
Next, we compare model responses on acceptability of familiar English Transfer or Manner of Accompanied Motion verbs in both the Dative and Ditransitive (or ``Double Object'' (DO)) construction.  We also test nonce verbs, which are established through context sentences as either Transfer or Manner of Accompanied Motion verbs, in both the Dative and Ditransitive.  We compare model responses to expected acceptability based on corpus studies and linguistic literature, from which we established the hypothesis that when context highlights a verb's Transfer semantics, the Ditransitive is not blocked, whereas when context highlights a verb's Manner of Accompanied Motion semantics, the Dative variant is acceptable whereas the Ditransitive variant would be conservatively avoided.  

Results for familiar Transfer and Manner of Accompanied Motion verbs are given in Table~\ref{tab:exp2cresults}.  We note that a random pick, demonstrating no preference for either verb class in either construction, would be 50\% pick rate for the Ditransitive/DO construction.  The expected preference is for Transfer verbs to pattern with the Ditransitive/DO construction about 50\% of the time (see second reference line of Table~\ref{tab:exp2cresults}), while Manner of Accompanied Motion verbs are never preferred in the Ditransitive/DO construction (0\% in the second reference line of Table~\ref{tab:exp2cresults}).  

We ask whether models replicate the following conservative generalization: while motion verbs readily occur in the Dative construction, Manner of Accompanied Motion verbs, including the familiar and novel verbs used in our evaluation, are unattested in the Ditransitive construction. Thus, the relevant blocking effect targets a semantic class rather than individual lexical items.  Our results show that the expected effect of blocking is demonstrated in model preference: models show a contrasting preference for Transfer verbs over Manner of Accompanied Motion verbs in the Ditransitive, but disprefer the Ditransitive overall. 

\begin{table*}[!h]
\centering
\resizebox{\textwidth}{!}{%
\begin{tabular}{llrrr}
\toprule
Tier & Model & Transfer DO\% & Motion DO\% & Class effect \\
\midrule
Reference & Random pick (no preference)   & 50\%      & 50\%     & \phantom{$+$}0 \\
Reference & Target (construction working) & $\sim$50\% & $\sim$0\% & $\sim+50$ \\
\midrule
Closed --- large       & GPT-5.4                & 37.5\% & 15.0\% & $+22.5$ \\
\addlinespace
Closed --- mid         & GPT-4o-mini            & 40.0\% & 10.0\% & $+30.0$ \\
Closed --- mid         & GPT-4o-mini (logprob)  & 40.0\% & 10.0\% & $+30.0$ \\
\addlinespace
Open --- MoE (13B active)       & DeepSeek V4-Flash      & 50.0\% & 50.0\% & \phantom{$+$}$+0.0$ \\
\addlinespace
Open --- large ($\sim$32B) & Qwen3-32B (instruct)     & 30.0\% & 20.0\% & $+10.0$ \\
\addlinespace
Open --- small (7--8B) & Qwen2.5-7B-Instruct    & 32.5\% & 15.0\% & $+17.5$ \\
Open --- small (7--8B) & Llama-3.1-8B base      & 25.0\% & \phantom{0}5.0\% & $+20.0$ \\
Open --- small (7--8B) & Llama-3.1-8B-Instruct  & 37.5\% & 27.5\% & $+10.0$ \\
\bottomrule
\end{tabular}
}
\caption{Double-Object (DO) construction preference rates for Transfer verbs vs.\ Manner of Accompanied Motion verbs, with the class effect (Transfer DO\% $-$ Motion DO\%). The expectation is that  Transfer verbs lead to a strong preference for DO while Motion verbs block DO.}
\label{tab:exp2cresults}
\end{table*}

Table~\ref{tab:exp2dresults} summarizes model responses when faced with nonce verbs where preceding context frames the nonce verb either as a Transfer or non-alternating Manner of Accompanied Motion verbs.  Similar to the results for a-adjective preemption, we see that most models no longer demonstrate a blocking or preference when facing nonce verbs. However, the effect is not as strong as what was witnessed with novel a-adjectives, as GPT-5.4 and GPT-4o-mini maintain a preference in the right direction.  This may be evidence that partial blocking is more recoverable from priors, perhaps because it is related to a semantic class, than fully idiosyncratic preemption.

\begin{table*}[!h]
\centering
\resizebox{\textwidth}{!}{%
\begin{tabular}{llrrr}
\toprule
Tier & Model & Transfer DO\% & Motion DO\% & Class effect \\
\midrule
Reference & Random pick (no preference)   & 50\%      & 50\%     & \phantom{$+$}0 \\
Reference & Target (construction working) & $\sim$50\% & $\sim$0\% & $\sim+50$ \\
\midrule
Closed --- large       & GPT-5.4                & 32.5\% & 20.0\% & $+12.5$ \\
\addlinespace
Closed --- mid         & GPT-4o-mini            & 35.0\% & 12.5\% & $+22.5$ \\
\addlinespace
Open --- MoE (13B active)       & DeepSeek V4-Flash      & 50.0\% & 50.0\% & \phantom{$+$}$+0.0$ \\
\addlinespace
Open --- large ($\sim$32B) & Qwen3-32B (instruct)    & \phantom{0}5.0\% & \phantom{0}0.0\% & \phantom{$+$}$+5.0$ \\
\addlinespace
Open --- small (7--8B) & Qwen2.5-7B-Instruct    & 30.0\% & 35.0\% & $-5.0$ \\
Open --- small (7--8B) & Llama-3.1-8B base      & \phantom{0}5.0\% & \phantom{0}0.0\% & \phantom{$+$}$+5.0$ \\
Open --- small (7--8B) & Llama-3.1-8B-Instruct  & 35.0\% & 32.5\% & \phantom{$+$}$+2.5$ \\
\bottomrule
\end{tabular}}
\caption{Double-Object construction preference rates for nonce Transfer verbs vs.\ nonce Manner of Accompanied Motion verbs, with the class effect (Transfer DO\% $-$ Motion DO\%).  }
\label{tab:exp2dresults}
\end{table*}

We may again ask if this is a failure of extending the conservative blocking, or a failure of recognizing the distinct Manner of Accompanied Motion and Transfer classes of verbs as such.  We therefore test models on accuracy in identifying the familiar and nonce verbs (where context cues either motion or transfer semantics) as either Manner of Accompanied Motion or Transfer verbs.  We show that all models are able to identify the classes accurately 100\% of the time (we report full results in Appendix A). The context-setting sentences establish the semantics of the nonce verbs: the models know exactly which novel verbs are Transfer and which are motion. The failure to demonstrate a preference for novel Transfer verbs within Ditransitive/DO is therefore a deployment failure, not an identification failure.  The model can classify the verb, it just does not extend the preemption pattern to the novel verbs.  
This mirrors our nonce a-adjective findings. In both experiments, the model recovers and identifies the relevant category (a-adjective class; transfer-vs-motion class) but does not productively deploy the preemption-based restriction on novel items. Identification is solved; productive deployment from the structure alone is the gap.

\section{Related Work}
\label{sec:rel-work}
The question of whether LLMs acquire grammatical knowledge beyond memorized surface patterns has been pursued from several directions. Our experiments sit at the intersection of three.

A first line of work probes LLMs for sensitivity to syntactic structure using targeted minimal pairs. This began with \citet{linzen2016assessing} on subject-verb agreement and was substantially extended by the BLiMP benchmark \cite{warstadt2020blimp}. These studies establish that models track structural dependencies well above chance, but they evaluate grammatical knowledge rather than productivity. They ask whether models can recognize a violation, not whether models can extend a pattern to novel material under the right contextual pressure. Our coercion experiment is closer in spirit to productivity probes: a model that has memorized \textit{drink the bottle} tells us little, but a model that interprets \textit{sip the glempkit} as targeting the contents has done generalization work.

A second line, more directly relevant, asks whether models encode construction-level knowledge in the sense of CxG. \citet{Tayyar_Madabushi_Romain_Divjak_Milin_2020} probe BERT for sensitivity to the form-meaning pairings of specific constructions, and later work demonstrates that model sensitivity to a construction depends, in part, upon its level of schematicity (whether it has fixed, phonological items or is comprised of fully flexible slots) \cite{bonial2025constructing}. \citet{weissweiler2022better,weissweiler2023explaining} examine the comparative correlative and a handful of other English constructions, finding that models recognize constructional forms but often fail on the corresponding meanings. \citet{scivetti2025beyond} also explore downstream understanding of constructional semantics and show that while models perform impressively on Natural Language Inference tasks involving common argument structure constructions, model performance degrades sharply on syntactically identical, less common constructions. \citet{mahowald2023discerning} tests GPT models on the \textit{a beautiful five days} construction and finds the pattern is reproduced, but with weaker semantic grounding than in humans. The general pattern these studies converge on is that models recognize constructional form and do less well on the semantics and the conditions on productive extension. This is consistent with what we find for the identification half of our preemption experiments. Our results provide some explanation for this trend.  The gap is not between form and meaning; it is between recognizing a category and withholding a construction from members of it.

A third line addresses overgeneralization and indirect negative evidence specifically. Most of this work focuses on children rather than models and we have already covered much of this research in \S\ref{sec:coercion} and \S\ref{sec:preemption}.  \citet{boyd2011learning}, \citet{robenalt2015judgment}, and \citet{perek2017linguistic} jointly establish that preemption and entrenchment shape how human learners retreat from overgeneralization, with the Dative alternation and a-adjectives as the canonical test cases. 

Whether LLMs deploy analogous mechanisms is comparatively underexplored. One notable exception is \citet{misra2024generating}, who examine LLM judgments on Dative alternation items with novel verbs. The novel verbs are presented in the Dative or Ditransitive, and information structure elements are adjusted such that the researchers test feature configurations, such as whether or not the recipient is realized as a pronoun or noun phrase, to find evidence for novel hypotheses about a novel verb's exposure conditions that facilitate or preempt Dative generalization.  Their finding is that models can successfully predict the alternation preferences of novel verbs that did not alternate in the training data. This is complementary to, rather than in conflict with, our result: in their setting, the novel verb is observed in a Dative construction, so the model generalizes from observed exposure, whereas our nonce verbs are never presented in either Dative variant, and any restriction must therefore be projected from the verb's context-established semantic class alone. Models succeed at the former; we find no evidence of the latter under this design.
While the present research tests a hypothesis gleaned from linguistic theory, the features that \citet{misra2024generating} test are derived in a bottom-up fashion from the data within models themselves, and they leave for future work testing these hypotheses with human subjects.

\section{Conclusions \& Future Work}
\label{sec:conclusions}
In this research, we have presented two opposing forces proposed in psycholinguistic and corpus studies.  Coercion facilitates creative productivity through a process of inferring the interpretation of unexpected lexical items within certain constructions.  Conversely, preemption operates to conservatively avoid certain novel combinations of constructions and lexical items, preventing overgeneralization of particular constructions.  

Our results indicate that models learn constructional types along with their lexical restrictions, allowing them to match human preferences in coercive contexts that trigger a metonymic construal of a container noun in which it denotes the contents of a container. Models  generalize from familiar linguistic information and commonsense knowledge to correctly interpret novel nouns as referring to either contents or containers, and thereafter infer and mirror the human-like preference for coercing containers to read as contents.  The finding is clear: models are able to generalize from positive evidence (mutual knowledge of the patterns of semantic and lexical-class selection that grammatical markers, words and constructions display) in reinterpreting novel nouns in coercive constructions.  

Our results also demonstrate that models mirror human preferences in conservatively avoiding certain familiar lexical items in the context of constructions where those lexical items have not been observed before.  Our second set of preemption experiments demonstrate that models can extend preemptive patterns \emph{but only when presented with familiar English test items}.  Models reflect human-like preemptive patterns blocking familiar a-adjectives in the pre-nominal position of the Attributive Modifier-Head construction.  Furthermore, models demonstrate partial blocking of familiar verbs with a Manner of Accompanied Motion construal in the Ditransitive/Double Object construction.  However, models do not extend preemptive, blocking patterns to novel lexical items of the same class.  Models do not appear to similarly generalize from negative evidence, from what is \textbf{not} said, in the case of statistical preemption.\footnote{Since the first version of this paper appeared (June 2026), \citet{wang2026disentangling} report converging findings. Using controlled rearing with periphrastic causative usages of intransitive-only verbs removed, they find that models avoid transitive overgeneralizations but show no verb-specific statistical preemption. Removing the preemptive evidence made the overgeneralized form less likely rather than more, indicating that models treat the competing construction as indirect positive evidence. They do report a weak abstract preemption effect when periphrastic causatives are removed across all verbs.}

In future work, we intend to target factors that might have depressed the use of the preemption strategy (e.g., discourse-context appropriateness).  We will shift our stimuli from Experiments 2 and 3 in order to test whether statistical preemption operates only when (a) the two alternative formulations are equally viable, and (b) the preempted structure is more attractive for other communicative reasons. This research will advance our understanding of how a context can strongly favor a particular construction (e.g., the Attributive, the Ditransitive) and yet fail to license it, owing to a competing form that preempts its use.

\textbf{We conclude that while models can generalize to a remarkable degree when this is based on positive frequency signals, current models do not deploy negative evidence in the way that people do.} Preemption requires learning from systematic \emph{absence}: the learner must compare forms with positive evidence against semantically equivalent alternatives that never occur, a comparison for which next-token prediction provides no direct training signal. 
Coercion is recoverable because it is top-down and semantic: the construction's lexical-class selectivity is evident in the context, and the model generalizes it. The conservative, preemptive signal is not present in the context or recoverable from constructional semantics---the incompatible lexical items are seemingly idiosyncratic.  This asymmetry helps to explain why models coerce but do not preempt.\footnote{We further speculate that we can attribute this difference to the fact that models are trained to interpret meaning in any and all of a user's potential utterances.  However, the model need not acquire the creative and conservative balance required to produce understandable, yet creative utterances that have never been seen in training data.}  Our research thus points to a critical generalization gap between models and people — one that clarifies where models succeed and fail, suggests targets for improving them, and informs theoretical accounts of the statistical forces shaping human language.

\bibliography{coercion-preemption-fixed}

\section*{Appendix A: Class Identification}
\label{app:classid}
Table~\ref{tab:nonce-categorization} shows models recognize even nonce a-adjectives as such close to 100\% of the time.  Table~\ref{tab:nonce-verb-categorization} demonstrates that models perfectly (100\% of the time) classify even nonce verbs as either Transfer or Manner of Accompanied Motion verbs based on the context-setting sentences. 

\begin{table}[!h]
\centering
\resizebox{\columnwidth}{!}{%
\begin{tabular}{lcc}
\toprule
Model & \shortstack{Nonce \emph{a-}adj $\to$\\``restricted'' (A)} & \shortstack{Nonce control $\to$\\``ordinary'' (B)} \\
\midrule
GPT-5.4           & \phantom{1}95\% (19/20)  & 100\% (20/20) \\
GPT-4o-mini       & 100\% (20/20) & \phantom{1}90\% (18/20)  \\
DeepSeek V4-Flash & 100\% (20/20) & 100\% (20/20) \\
\bottomrule
\end{tabular}}
\caption{Categorization accuracy on nonce items: rate at which models classified nonce \emph{a-}adjectives as ``restricted'' and nonce controls as ``ordinary.''}
\label{tab:nonce-categorization}
\end{table}

\begin{table}[!h]
\centering
\resizebox{\columnwidth}{!}{%
\begin{tabular}{lcc}
\toprule
Model & Transfer $\to$ ``transfer'' & Motion $\to$ ``motion'' \\
\midrule
GPT-5.4           & 100\% (20/20) & 100\% (20/20) \\
GPT-4o-mini       & 100\% (20/20) & 100\% (20/20) \\
DeepSeek V4-Flash & 100\% (20/20) & 100\% (20/20) \\
\bottomrule
\end{tabular}}
\caption{Categorization accuracy on nonce verbs: rate at which models classified transfer-of-possession nonce verbs as ``transfer'' and Manner of Accompanied Motion nonce verbs as ``motion.''}
\label{tab:nonce-verb-categorization}
\end{table}

\end{document}